\documentclass[journal]{IEEEtai}

\usepackage{graphicx} 
\usepackage{url}
\usepackage{tabularx}
\usepackage{color}
\usepackage{tabularray}
\definecolor{Shark}{rgb}{0.129,0.145,0.16}
\begin{document}
\title{A Survey on Generative AI and LLM for Video Generation, Understanding, and Streaming}

\author{Pengyuan Zhou, Lin Wang, Zhi Liu, Yanbin Hao, Pan Hui, Sasu Tarkoma, Jussi Kangasharju
\thanks{Corresponding author: Pengyuan Zhou (zpymyyn@gmail.com)}
}

\maketitle

\begin{abstract}
This paper offers an insightful examination of how currently top-trending AI technologies, i.e., generative artificial intelligence (Generative AI) and large language models (LLMs), are reshaping the field of video technology, including video generation, understanding, and streaming. It highlights the innovative use of these technologies in producing highly realistic videos, a significant leap in bridging the gap between real-world dynamics and digital creation. The study also delves into the advanced capabilities of LLMs in video understanding, demonstrating their effectiveness in extracting meaningful information from visual content, thereby enhancing our interaction with videos. In the realm of video streaming, the paper discusses how LLMs contribute to more efficient and user-centric streaming experiences, adapting content delivery to individual viewer preferences. This comprehensive review navigates through the current achievements, ongoing challenges, and future possibilities of applying Generative AI and LLMs to video-related tasks, underscoring the immense potential these technologies hold for advancing the field of video technology related to multimedia, networking, and AI communities.
\end{abstract}
\begin{IEEEImpStatement}
This paper contributes to the field of video technology by examining the integration of Generative AI and Large Language Models (LLMs) in video generation, understanding, and streaming. Its exploration of these technologies offers a foundational understanding of their potential and limitations in enhancing the realism and interactivity of video content. The exploration of LLMs in video comprehension sets the stage for advancements in accessibility and interaction, promising enhanced educational tools, improved user interfaces, and advanced video analytics applications. Additionally, the paper underscores the role of LLMs in optimizing video streaming services, leading to more personalized and bandwidth-efficient platforms. This could substantially benefit the entertainment sector with adaptive streaming solutions tailored to individual preferences. By identifying key challenges and future research directions, the paper guides ongoing efforts to merge AI with video technology, while raising awareness about potential ethical issues. Its influence extends beyond academia, encouraging responsible AI development and policy-making in video technology, balancing technological advancements with ethical considerations.
\end{IEEEImpStatement}
\begin{IEEEkeywords}
Generative Artificial Intelligence (AI), Large Language Model (LLM), Video Understanding, Video Generation, Video Streaming, GPT
\end{IEEEkeywords}

\section{Introduction}
The creation, analysis, and delivery of video content have all undergone significant breakthroughs in recent years thanks to exciting advancements in video-related technology. Academia and industry have worked to push the limits of what is feasible in the field of video processing, from creating realistic videos to comprehending complicated visual environments and optimizing video streaming for better user experiences. Integrating \textbf{Generative AI} and \textbf{LLM} can open up exciting possibilities in video-related fields.

With the ability to create lifelike and contextually consistent videos, video creation has emerged as an intriguing study field. Researchers have made significant progress in producing movie clips that reveal fine details and capture the essence of real-world dynamics by utilizing deep learning methods such as Generative Adversarial Networks (GANs). However, challenges such as long-term video synthesis consistency and fine-grained control over created content are still under exploration.

Similar developments have been made in video understanding, which entails gleaning important information from video clips. Conventional techniques depend on manually created features and explicit modeling of video dynamics. Recent advancements in language and vision have made considerable progress. Pre-trained transformer-based architectures, like OpenAI's GPT, among other LLMs, in general, have shown impressive talents in processing and producing textual data. These LLMs hold great potential for video-understanding tasks like captioning, action identification, and temporal localization.

Furthermore, improving video delivery has become increasingly important and challenging due to the rising demand for high-quality, high-resolution, and low-latency video service demands. Offering seamless and immersive streaming experiences is significantly hampered by bandwidth restrictions, network jitters, and different user preferences. By providing context-aware video distribution, real-time video quality improvement, and adaptive streaming depending on user preferences, LLMs present an exciting approach to overcoming these difficulties.

In light of these advancements, this study thoroughly analyzes the potential of Generative AI and LLMs for generating, comprehending, and streaming videos. We review existing works to try to answer the following questions:
\begin{itemize}
    \item What technologies have been proposed and are revolutionizing the mentioned video research fields?
    \item What technical challenges remain to be addressed to push forward the use of GAI- and LLM- methods in the mentioned video services? 
    \item What unique concerns have been raised due to the employment of GAI and LLM methods? 
\end{itemize}
We hope to draw attention from the multimedia, networking, and AI communities to encourage future research in this fascinating and quickly developing area. 

\begin{table*}[t]
\centering
\setlength{\tabcolsep}{0.5em} 
{\renewcommand{\arraystretch}{1.6}
\begin{tabularx}{\textwidth}{|l|c|c|c|c|c|X|}
\hline
\textbf{Year} & \textbf{GenAI} &  \textbf{LLM} &  \textbf{Generation} &  \textbf{Understanding} &  \textbf{Streaming}  & \textbf{Summary} \\
\hline
\hline
\cite{bhagwatkar2020review}, 2020 &  $\surd$ & X & $\surd$ & X & X  & Overview of VAEs, GANs, and Transformers for video generation. \\
\hline
\cite{singh2023survey}, 2023 & $\surd$ & X & $\surd$ & X & X &  Investigates Text-to-Image and Text-to-Video AI generators. \\
\hline
\cite{liu2023ai}, 2023 & $\surd$ & X & $\surd$ & X & X  & Focus on AI methods for generating persuasive videos. \\
\hline
\cite{aldausari2022video}, 2022 &$\surd$ & X & $\surd$ & X & X & Focus on GAN methods for video generation.\\
\hline
\cite{rafiq2023video}, 2023 &  X & X & X & $\surd$ & X &   Focus on deep learning methods for description.\\
\hline
\cite{singh2020comprehensive}, 2020  & X & X & X & $\surd$ & X & Survey description methods for specific datasets.\\
\hline
\cite{aafaq2019video}, 2019  & X & X & X & $\surd$ & X &  Methods, datasets and metrics for AI-based video description.\\
\hline
Ours, 2023 & $\surd$ & $\surd$ &$\surd$ &$\surd$ &$\surd$ &GenAI and LLM for video generation, understanding, and streaming.\\
\hline
\end{tabularx}
}
\caption{Relevant survey papers in recent years.}
\label{table:surveys}
\end{table*}

\section{Methodology}
This survey targets a wide view of the interaction between Generative AI and LLMs and the video field. It covers more than 100 papers collected from Google Scholar, IEEE Xplore, ACM Digital Library, Elsevier, ScienceDirect, DBLP, etc. The queries combine the following keywords: Generative AI / LLM $\&$ Video Understanding / Segmentation / Generation / Streaming, and the keywords related to the key technologies discussed in Section~\ref{sec:framework}. We further complement the articles by adding prominent research featured on the Internet to cover a comprehensive set of important publications in this area. This process was continued until no new articles were found. We have carefully examined the papers and selected the most relevant and important articles while filtering out the less relevant ones. The selected papers form the core of this survey, and we have performed continuous updates during the survey writing process to cover papers published since the start of our process. Note that due to the rapid development and large number of publications in relevant fields in 2023, there might be some good new papers we overlooked; however, we have made our best efforts.

\section{Overview}\label{sec:framework}
We envision Generative AI and LLMs play key roles in the full life cycle of video, from generation, and understanding, to streaming. The framework crosses three major computer science communities, i.e., AI, Multimedia, and Networking. AI community is witnessing an unprecedented development rate that takes only roughly a year from models capable of text-to-image generation to those capable of text-to-video generation, from 2021 to 2022. Now there are even demos showing the ability to create 3D videos just by using prompts. Therefore, we can only imagine Generative AI becoming more important for the video generation industry, outrunning or even replacing entirely the conventional generation methodologies. Video understanding is useful for many cases, e.g., scene segmentation, activity monitoring, event detection, and, video captioning, a rising direction that gets increasing attention. Since 2023, the LLMs' capabilities of understanding multimodal input such as images and videos have also been significantly promoted by the most advanced products like GPT-4 and Video-ChatGPT~\cite{Maaz2023VideoChatGPT}. As for video streaming, LLMs also hold interesting potential to improve several key steps of the streaming pipeline. For instance, a model with improved understanding capability can grasp the semantic meaning of the video scenes and optimize the transmission by varying the encoding rate accordingly. Further, 3D video streaming such as point cloud which is widely used in XR games, can benefit from LLM's understanding of the surroundings to predict the user's FoV in the next moment to conduct content pre-fetching.

\subsection{Main Components}
The synergy between Generative AI and LLMs has opened new frontiers in video generation, crafting visuals that are increasingly indistinguishable from reality. These technologies work together to enrich the digital landscape with innovative content as follows (Section~\ref{subsec:generation}):
\begin{itemize}
  \item GANs (Generative Adversarial Networks) leverage the creative adversarial process between generative and discriminative networks to understand and replicate complex patterns, resulting in lifelike video samples.
  \item VAEs (Variational Autoencoders) generate cohesive video sequences, providing a structured probabilistic framework for the seamless blending of frames that narratively make sense.
  \item Autoregressive models create sequences where each video frame logically follows from the last, ensuring a narrative and visual continuity that captivates viewers.
  \item Diffusion models convert intricate textual narratives into detailed and high-resolution videos, pushing the boundaries of text-to-video synthesis.
\end{itemize}

Next, LLMs enhance video comprehension by providing contextually rich interpretations and descriptions, facilitating a deeper engagement with video content (Section~\ref{subsec:understanding}):
\begin{itemize}
  \item Video captioning employs LLMs to generate insightful and accurate descriptions, capturing the essence of the visual content in natural language, making videos more searchable and accessible.
  \item Video question answering harnesses the contextual understanding capabilities of LLMs to field complex viewer inquiries, providing responses that add value and depth to the viewing experience.
  \item Video retrieval and segmentation are revolutionized by LLMs, which parse and categorize video content into intelligible segments, streamlining the searchability and navigation of extensive video libraries.
\end{itemize}
\begin{figure*}[htb!]
	\centering
	\includegraphics[width=0.8\textwidth]{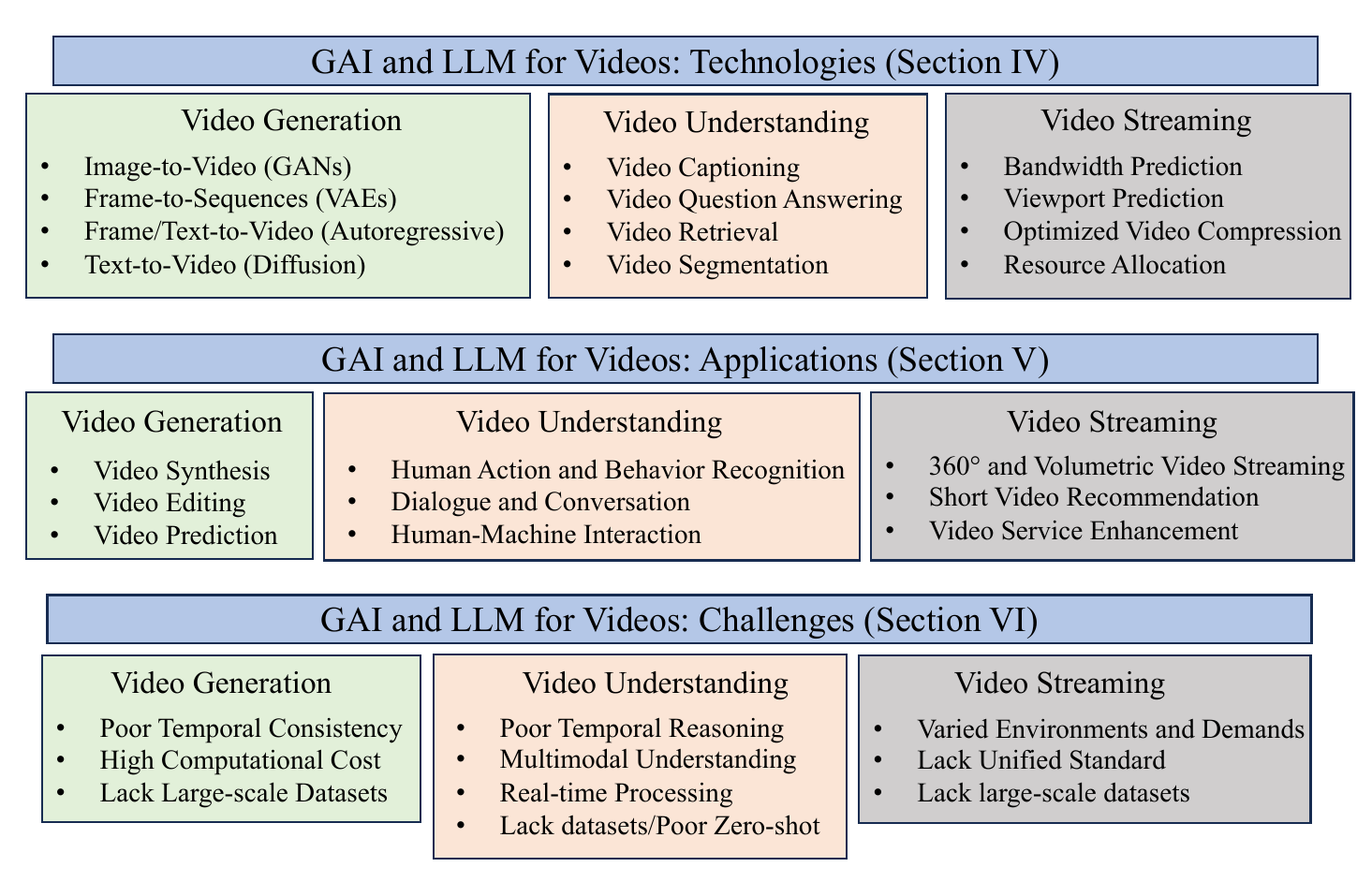}
	\caption{Taxonomy of video generation, understanding, and streaming with GAI and LLMs.}
	\label{fig:framework}
\end{figure*}
Last but not least, LLMs can redefine the streaming landscape by optimizing bandwidth usage, personalizing content delivery, and enhancing viewer interaction from the following perspectives (Section~\ref{subsec:streaming}):
\begin{itemize}
  \item Bandwidth prediction is refined through LLMs that analyze past and present network data, predicting future demands to allocate resources proactively, thereby ensuring uninterrupted streaming.
  \item Viewpoint prediction is augmented by LLMs' comprehension of content and user behavior, anticipating the next focus area within a video to deliver a tailored and immersive viewing experience.
  \item Video recommendation and resource allocation are advanced by the analytical prowess of LLMs, matching viewer preferences with content and managing network resources to deliver a customized and efficient streaming service.
\end{itemize}

\section{Technologies}

\subsection{Generative AI for Video Content Generation}\label{subsec:generation} 
Generative AI has emerged as a powerful tool for creating a wide range of content, including images, text, music, and video. For video content creation, generative models have the potential to revolutionize the way we create and consume video by automating the generation of realistic and high-quality content. 
Generative models, especially deep learning-based generative models such as GANs~\cite{vondrick2016generating}, Variational Autoencoders (VAEs)~\cite{denton2018stochastic}, autoregressive models~\cite{kalchbrenner2017video}, and diffusion-based models~\cite{ho2022video,ho2022imagen,singer2022make}, have demonstrated remarkable success in generating realistic and diverse content in various domains. These models learn the underlying data distribution by training on large datasets, enabling the generation of samples that resemble the training data. Some of the SOTA generative models are listed in table \ref{reviewed_gen_methods}. However, generative AI models face unique challenges in the context of video content generation due to the spatial-temporal property of videos, the requirement of photo-realistic dynamic scenes, and the considerable cost of processing video data. 
Despite these challenges, significant progress has been made in developing generative models for video content creation. We now discuss them in detail.

\begin{figure*}[t!]
	\centering
	\includegraphics[width=0.8\textwidth]{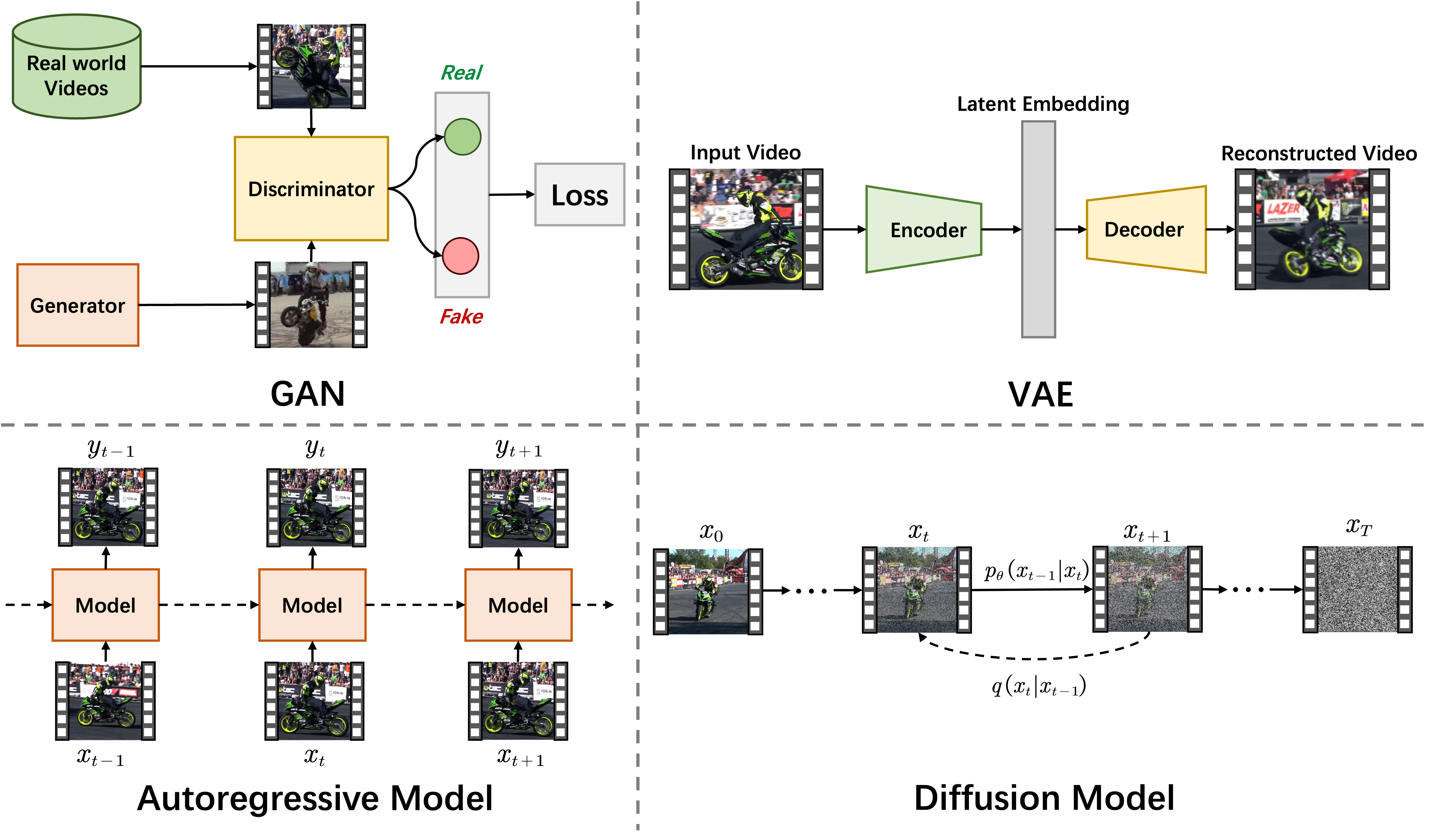}
	\caption{An overview of advanced AI-based video generation technologies.}
	\label{fig:generationtech}
\end{figure*}

\begin{table*}[htbp]
\centering
\caption{The reviewed generative methods for video content generation.}
\begin{tabular}{l|l|l}
\hline
Method &Input information &Task \\
\hline
 \multicolumn{2}{c}{\textbf{\textit{GAN models}}} & \\
\hline
VideoGAN~\cite{vondrick2016generating} & Video & Video generation and video prediction given a single static image in a close-set scene domain. \\
EDN~\cite{chan2019everybody} & Video &  Video-to-video translation using pose as an intermediate representation.  \\
\hline
\multicolumn{2}{c}{\textbf{\textit{VAE models}}} & \\
\hline
SVG~\cite{denton2018stochastic}  & Video & Video prediction given the initial frames of a simple motion video like human activity \\
SadTalker~\cite{zhang2023sadtalker} &Image, audio & Talking head generation given a face image and a piece of speech audio. \\
\hline
\multicolumn{2}{c}{\textbf{\textit{Autoregressive models}}} & \\
\hline
Video Pixel Networks~\cite{kalchbrenner2017video} & Video & Video prediction~given the initial frames of a simple motion video like MNIST motion. \\
CogVideo~\cite{hong2022cogvideo}   & video, text       & Text-to-video generation, video prediction, and video frame interpolation. \\
\hline
\multicolumn{2}{c}{\textbf{\textit{Diffusion models}}} & \\
\hline
VDM~\cite{ho2022video} & Video, text/label & Text-conditioned or label-conditioned video generation, and video prediction. \\
Imagen-Video~\cite{ho2022imagen} & video, text & Text-to-video generation, video prediction, and video frame interpolation.  \\
Make-a-Video~\cite{singer2022make} & video, text &  Text-to-video generation, video prediction, and video frame interpolation.\\
Video LDM~\cite{blattmann2023align} & video, text &  Text-to-video generation, high-Resolution video synthesis. \\
DreamTalk~\cite{ma2023dreamtalk}    & Image, audio  &  Talking head generation given a face image and a piece of speech audio.\\
Dancing Avatar~\cite{qin2023dancing} & Motion, text &  Generating highquality human videos guided by textual descriptions and
motion. \\
Discro~\cite{wang2023disco}   & Motion, text &  Generating highquality human videos guided by textual descriptions and
motion. \\
\hline
\end{tabular} 
\label{reviewed_gen_methods}
\end{table*}

\noindent\textbf{GANs} consist of a generator and a discriminator, which are trained in a two-player min-max game. The generator learns to generate realistic samples, while the discriminator learns to distinguish between generated samples (i.e., fake) and ground truth (GT) samples (i.e., real). For video generation, GANs have been extended to model temporal consistency and generate realistic video frames. An example is VideoGAN~\cite{vondrick2016generating}, introducing a two-stream architecture to separately model appearance and motion in videos. The generator produces video frames, while the discriminator evaluates the realism of individual frames and the motion between consecutive frames. This approach is successful in generating realistic videos of human actions and scenes.

\noindent\textbf{Variational Autoencoders (VAEs)} are generative models that learn a probabilistic mapping between the data space and a latent space by optimizing a variational lower bound on the data likelihood. In the context of video generation, VAEs have been adapted to model the temporal structure of videos and generate video sequences. One example is the Stochastic Video Generation (SVG) framework~\cite{denton2018stochastic}, which extends VAEs to model the distribution of future video frames conditioned on past frames. The SVG framework introduces a hierarchy of latent variables to capture the multi-scale nature of video data, enabling the generation of diverse and realistic video sequences.

\noindent\textbf{Autoregressive models} generate data by modeling the conditional distribution of each data point given its preceding data points. In the context of video generation, autoregressive models can be used to generate video frames sequentially, conditioning each frame on the previously generated frames. A prominent example is the Video Pixel Networks ~\cite{kalchbrenner2017video}, an autoregressive model that extends the PixelCNN~\cite{van2016conditional} to model video data. VPN encodes video as a four-dimensional dependency chain, where the temporal dependency is captured using an LSTM and the space and color dependencies are captured using PixelCNN. Transformer~\cite{vaswani2017attention}, on the other hand, models the sequential data and performs well at many NLP and vision tasks. In contrast to GAN-based methods, autoregressive models can deal with both continuous and discrete data.

\noindent\textbf{Diffusion models} construct data generation as a denoising process. DMs have recently shown remarkable success in visual generation and achieved a notable state-of-the-art performance on most image-related synthesis or editing tasks. Video diffusion model (VDM)~\cite{ho2022video} is the first work that introduces DMs to the domain of video generation by extending the U-net~\cite{ronneberger2015u} to a 3D version. Later, Imagen-Video~\cite{ho2022imagen}, by taking the merit of its strong pretrained text-image generator Imagen, exhibits substantial capability in high-resolution text-video synthesis. It interplaces the temporal attention layer in serial spatial layers to capture the motion information. Make-a-Video~\cite{singer2022make} is another powerful competitor in text-video synthesis by conditioning on the CLIP~\cite{radford2021learning} semantic space. It first generates the keyframes conditioning text prior information and then cascaded with several interpolations and upsampling diffusion model to achieve high consistency and fidelity. However, both aforementioned pioneer works suffer from high computational costs, and Video LDM~\cite{blattmann2023align} is proposed to alleviate the problem by generating motion-aware latent representations in a semantically compressed space.

\subsection{LLMs for Video Scene Understanding}\label{subsec:understanding} 
Video scene understanding is a task that aims to extract meaningful information from videos. It involves recognizing objects, activities, and events in a video and understanding the relationships between them~\cite{chang2023survey}. Generative AI and LLMs have emerged as promising approaches for video scene understanding due to their ability to learn from large amounts of data and generate natural language descriptions of video contents~\cite{chen2023videollm}. In this paper, we discuss the use of LLMs for video scene understanding and review some of the techniques that have been proposed in recent years.

Video scene understanding involves several subtasks, including object detection, action recognition, and event detection~\cite{zhu2020comprehensive}. Object detection aims to identify and localize objects in a video, while action recognition aims to recognize human actions such as walking, running, and jumping. Event detection aims to identify and classify events such as accidents, sports events, and concerts. These sub-tasks are challenging because videos are complex and dynamic, and the same object or action can appear in different ways and contexts.

LLMs are neural network models that are trained on large amounts of text data to generate natural language text. These models have achieved impressive results in natural language processing tasks such as language translation, question-answering, and text generation. LLMs can also be used for video scene understanding by generating natural language descriptions of the video content~\cite{chen2023videollm}. These descriptions can help to summarize the video content and provide insights into the objects, actions, and events in the video.

\begin{figure}[t!]
    \centering
    \includegraphics[width=\columnwidth]{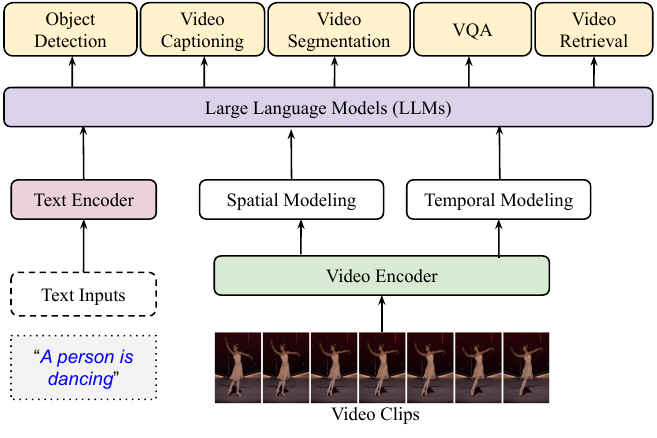}
    \caption{An overview of LLMs for video scene understanding tasks.}
    \label{fig:video-llms}
\end{figure}

Several methods have been proposed for using LLMs for different tasks in video scene understanding. Although different tasks hold different requirements regarding the way to use LLMs, we find they share some common components, such as temporal and semantic feature extraction from video clips, semantic and video feature alignment, \textit{etc.},  as illustrated in Fig.~\ref{fig:video-llms}. In the following, we discuss some of these techniques and their advantages and limitations.

\noindent \textbf{Video Captioning}
is a task that involves generating natural language descriptions of the video content~\cite{bain2023understanding,wu2023next}. This task can be approached using LLMs by training them on a large dataset of videos with corresponding captions. It involves two major steps. Firstly, the extracted visual and audio features are encoded into a fixed-length vector representation using the trained LLM~\cite{lai2023lisa,yang2023vid2seq}. This encoding captures the essential information from the video and provides contextual cues for generating accurate captions. Then, the LLM generates textual descriptions or captions for the video. These captions can encompass a range of details, including objects, actions, events, or any other relevant information that describes the video content effectively~\cite{zhao2023learning,dave2022hierarchical}.

Video captioning using LLMs finds application in various areas, including enhancing accessibility for individuals with hearing impairments, facilitating video search and retrieval, generating video summaries, and improving overall understanding of video content~\cite{ma2023llavilo}.

\noindent \textbf{Video Question Answering}
is a task that involves answering natural language questions about the video content. This task can be approached using LLMs by training them on a large dataset of videos with corresponding questions and answers~\cite{shao2023prompting,de2023visual,shao2023prompting}. The model learns to extract relevant information from the video content to answer the question. The advantage of this approach is that it can generate specific answers to specific questions. However, the limitations of this approach are that it requires large amounts of labeled data and it may not capture the context and complexity of the video content~\cite{singh2023visual,guo2023images,salaberria2023image}.

\noindent \textbf{Video Retrieval}
using LLMs refers to the process of searching and retrieving relevant videos from a large video database using advanced language models. LLMs are powerful neural network models that can understand and generate human-like text based on large amounts of training data~\cite{liu2023one,ma2023llavilo}. This task can be approached using LLMs by training them on a large dataset of videos with corresponding textual descriptions.
The representative approaches~\cite{zhao2023learning,maaz2023video} learn to associate the visual content of the video with the corresponding textual description, as depicted in Fig.~\ref{fig:video-llms}. With the power of LLMs, it enables more accurate and efficient video retrieval, improving the user experience and enhancing the utility of video databases. However, the limitations of this approach are that it requires large amounts of labeled data and may not capture the fine-grained details of the video content~\cite{hu2023reveal,yuksekgonul2022and}.

\noindent \textbf{Video Segmentation}, the task of segmenting objects or regions of interest in videos, can benefit from the application of LLMs~\cite{gao2023deep}.  LLMs can aid in semantic video segmentation by leveraging their language understanding capabilities. By incorporating textual descriptions or prompts, LLMs can guide the segmentation process, providing high-level context and semantic understanding. For instance, LLMs can generate textual masks or descriptions that describe the desired object or region to be segmented, assisting in accurate and contextually relevant segmentation~\cite{lai2023lisa,zhang2023video}. 
Moreover, video segmentation often requires temporal reasoning to accurately segment objects or regions over time. LLMs can be utilized to model long-range temporal dependencies and capture contextual information across video frames. By incorporating temporal cues into the language prompts or training LLMs with temporal objectives, they can facilitate temporal video segmentation, allowing for more coherent and consistent segmentations~\cite{ma2023llavilo}.

In a nutshell, LLMs have emerged as a promising approach for video scene understanding due to their ability to learn from large amounts of data and generate natural language descriptions of the video content. The techniques discussed in this paper demonstrate the potential of LLMs for video scene understanding. However, these techniques also have limitations, such as the requirement for large amounts of labeled data and the inability to capture fine-grained details of the video content. Further research is needed to improve the performance of LLMs for video scene understanding and to overcome these limitations.

\subsection{LLM for Video Streaming}\label{subsec:streaming} 
\begin{figure*}[htb!]
	\centering
	\includegraphics[width=0.65\textwidth]{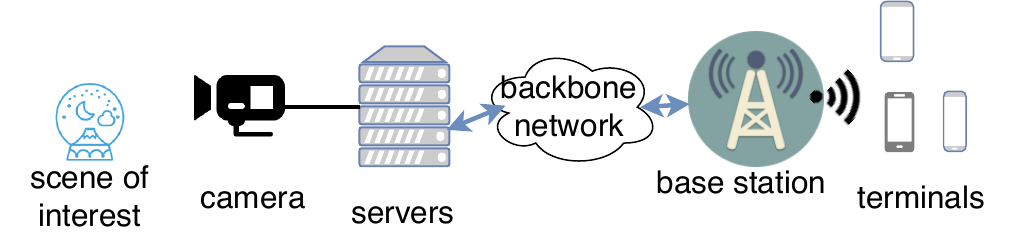}
	\caption{Illustration of a typical video transmission system. The scene of interest is captured by multiple cameras and the compressed video is conveyed to servers. The videos are distributed through the backbone network and directly received by mobile users from the corresponding wireless base station. }
	\label{fig:systemsingle}
\end{figure*}

Next, we explore how ChatGPT-like LLMs can enhance the video streaming experience from various perspectives. As illustrated in Fig.~\ref{fig:systemsingle}\footnote{Note that occasionally only part of this system is considered in a specific work.}, a typical video system consists of video capturing, video encoding (i.e., compression), video network transmission, video decoding, and video frame recovery. We first discuss the trending video formats with their featuring challenges. Then we summarize the potential of LLM for video streaming to tackle the challenges.

\noindent\textbf{LLMs for Bandwidth Prediction.} The future bandwidth prediction is a fundamental issue for improving video transmission. Bandwidth data is temporal; currently, a significant amount of work relies on deep learning methods like LSTM and RNN. Large-scale forecasting models can offer substantial advantages in predicting time series, enabling better anticipation of future network conditions and serving as a cornerstone for video transmission. Moreover, in new environments where sample scarcity is a concern, effective utilization of LLMs and transfer learning techniques can produce promising results even with limited samples. For example, Azmin et al. \cite{azmin2022bandwidth} presented a transformer-based model designed for 5G datasets, demonstrating significant enhancements compared to schemes relying solely on LSTM. They introduced novel feature analysis techniques, including LASSO and Random Forest with updated hyper-parameters, alongside the existing Random Forest with Informer.

\noindent\textbf{LLMs for Viewport Prediction.} 
One critical aspect of the VR/360° and other immersive video systems is viewport prediction, which involves accurately anticipating the user's next viewpoint within the virtual environment \cite{li2023spherical,van2022machine}. This prediction is crucial for ensuring a seamless and responsive viewing experience. To enhance viewport prediction, we can leverage the capabilities of LLMs like the GPT-4, which have shown exceptional performance in NLP and generation tasks. By adapting such language models to handle video-related data, we can significantly improve view angle prediction accuracy. The process involves training the LLM on vast datasets containing video sequences, user interaction patterns, and positional data to learn complex patterns and dependencies in user behavior, resulting in better predictions for the user's next view angle. For instance, the work by \cite{chao2021transformer} introduces a transformer-based approach for predicting viewports in 360° videos. This technique focuses solely on analyzing past viewport scanpaths to achieve precise long-term viewport predictions while maintaining low computational complexity. In the study conducted by \cite{cheng2022gaze}, transformers are incorporated to evaluate their efficacy in gaze estimation. By retaining convolutional layers and combining CNNs with transformers, the transformer functions as a complementary component to enhance the overall performance of CNNs, resulting in excellent performance. Additionally, \cite{ni2023human} combines gaze features with scene contexts and the visual characteristics of human–object pairs, through a spatiotemporal transformer to forecast human–object interactions in videos.

\noindent\textbf{Optimized Video Compression.}
LLMs can optimize video coding and compression, reducing file sizes and improving transmission efficiency. 
For instance, \cite{xiang2022mimt} put forward a masked image modeling transformer designed for deep video compression. Following the concept of a proxy task in pretrained language/image models, the transformer undergoes training to fully exploit temporal correlation among frames and spatial tokens within a few autoregressive steps. Meanwhile, \cite{mentzer2022vct} introduced a transformer-based approach to neural video compression that is elegantly simple, surpassing previous methods in performance without relying on architectural priors such as explicit motion prediction or warping.

\noindent\textbf{Resource Allocation.}
In wireless communication networks, resource allocation is a critical task that involves efficiently distributing limited network resources such as bandwidth, power, and time slots among various users and applications. Video streaming, being one of the most data-intensive and popular applications, demands careful resource allocation to ensure smooth and high-quality video delivery to users.

LLMs can process and analyze various textual inputs related to video streaming, including user preferences, video content descriptions, network conditions, and other contextual data. Using this information, LLMs can better understand user demands, video characteristics, and network requirements, to propose optimized resource allocation strategies. These strategies aim to prioritize and allocate resources in a way that maximizes the quality of video streaming, minimizes buffering or latency issues, and enhances the overall user experience.

Moreover, LLMs can continuously learn from vast amounts of data, adapting their resource allocation decisions over time based on changing network conditions and user behavior. This adaptability allows the resource allocation process to be dynamic and responsive to real-time changes, leading to more efficient and adaptive video streaming services.

\section{Applications}
\subsection{Generation}
\begin{figure}[t!]
	\centering
	\includegraphics[width=0.95\columnwidth]{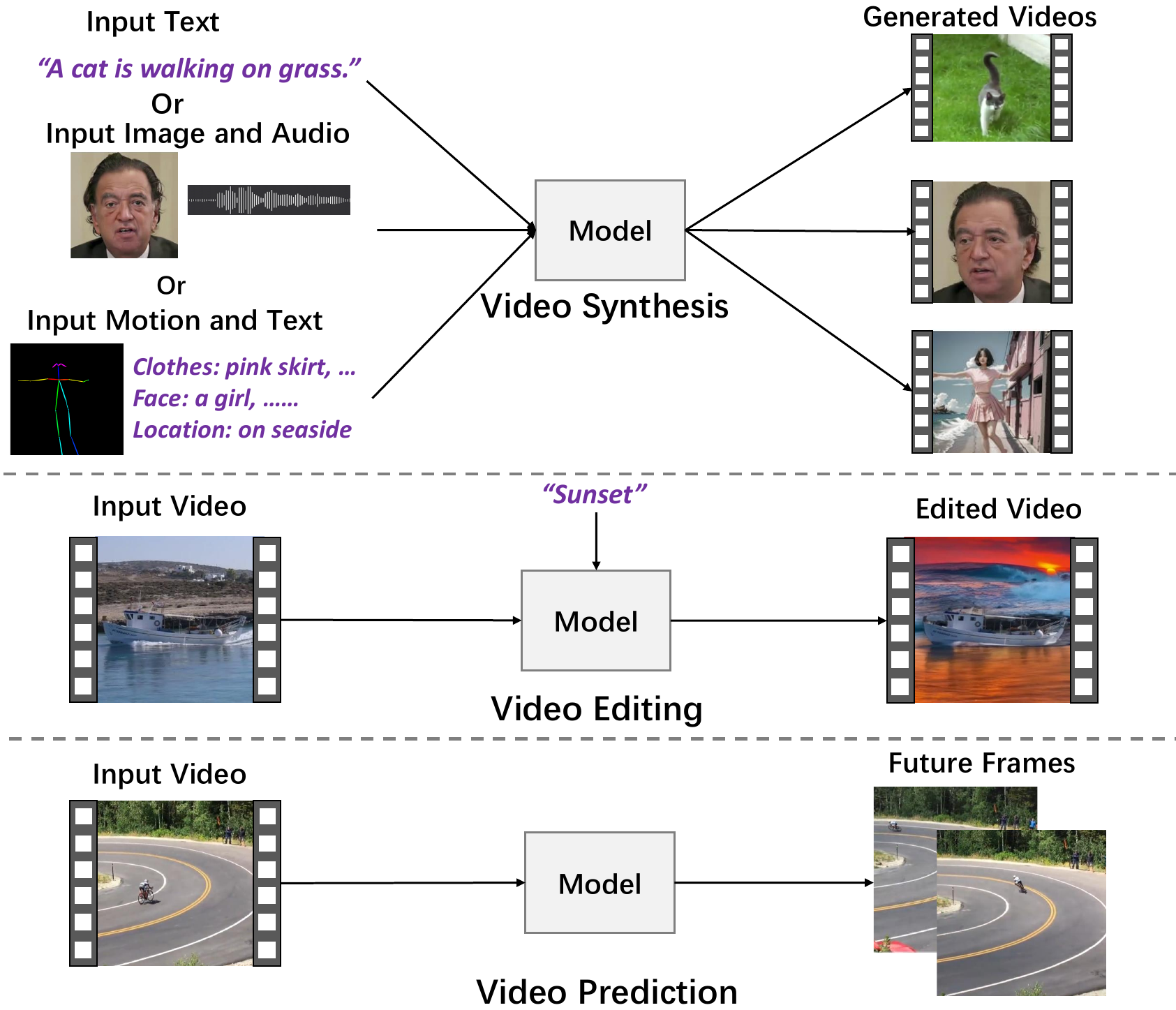}
	\caption{Video Generation Applications.}
	\label{fig:generationapplication}
\end{figure}

\noindent \textbf{Video Synthesis}. Generative AI models can be used to synthesize novel video content, enabling the creation of realistic scenes and special effects without manual intervention. Due to the inherent training instability of GANs, relatively fewer explorations have been conducted on GAN-based models for cross-modal video synthesis. TGAN~\cite{ding2019tgan}, as an early attempt, utilizes GAN for video generation by first generating a latent representation using a temporal generator and decoding it into pixels using an image generator. NUWA~\cite{wu2022nuwa}, a transformer-based model, proposes a unified cross-modal generative model capable of accommodating various generative scenarios such as text-to-video, sketch-to-video, video prediction, and more. CogVideo~\cite{hong2022cogvideo} extends the text-to-image model CogView~\cite{ding2021cogview} by implementing a multi-frame-rate hierarchical training strategy to better align the text and video clips. Recent diffusion-based models such as Imagen-Video~\cite{ho2022imagen} and Make-a-Video~\cite{singer2022make} have pushed the boundaries of video generation to a new level. However, these diffusion models suffer from a large number of parameters and complex cascaded networks, which greatly limit the community's ability to develop them further. Compared to other approaches, Video LDM~\cite{blattmann2023align} exhibits both efficiency and expressiveness. It achieves this by fine-tuning the publicly available Stable Diffusion (SD) Image LDM model using a vast dataset of 10.7 million video-caption pairs from the WebVid dataset~\cite{bain2021frozen}. Text2Video-Zero~\cite{khachatryan2023text2video} takes this a step further by proposing a method that does not rely on video data. Instead, it employs pre-defined global translation parameters to warp the latent code and utilizes cross-attention with the start frame to obtain consistent and denoised frames. Video LDM and Text2Video-Zero have also emerged with the capability of personalized video generation. Users can customize the concepts within the video using methods like Dreambooth~\cite{ruiz2023dreambooth}.

There are also works for domain-specific video synthesis tasks, such as audio-based video generation and human dancing video generation~\cite{qin2023dancing}. SadTalker~\cite{zhang2023sadtalker} leverages a conditional VAE to synthesize head motion and realize stylized audio-driven talking face animation. DreamTalk~\cite{ma2023dreamtalk} utilizes a diffusion model to generate highly diverse talking heads based on the provided source audio or video. For human dancing video generation, the GAN-based pose-guided video generation model, EDN~\cite{chan2019everybody}, is fine-tuned on image-pose pairs extracted from a specific human dancing video. It is capable of generating a person's image conditioned on any open-set pose image. However, EDN faces challenges in efficiently and accurately reconstructing human attribute details without extensive pre-training. Discro~\cite{wang2023disco} addresses this issue by leveraging the current state-of-the-art pre-trained diffusion model and a structural conditioning technique. To enhance attribute details during inference, it employs Grounded-SAM~\cite{kirillov2023segment} for foreground extraction and pre-trains the model on an extensive human-attribute dataset, achieving improved compositional aspects in dance synthesis.


Another line of research focuses on enhancing the smoothness of text-guided video generation by integrating current Large Language Models (LLMs). In order to better align visual tokenization with the learning process of LLMs, MAGVIT-v2~\cite{yu2023language} is proposed as a concise and expressive video tokenizer. This enables improved video generation performance of LLMs compared to diffusion-based models. VideoPoet~\cite{kondratyuk2023videopoet}, functioning as a versatile video generation model, utilizes a range of modal input tokenizers, including MAGVIT-v2, to facilitate video tokenization. It is capable of handling various video generation scenarios, involving the seamless conversion between video and other modalities such as text and audio.

\noindent \textbf{Video Editing} allows users to customize edits for a given video. Such applications are not limited to the capabilities of limited synthesis models, allowing the model to focus on editing specific scenes for improved temporal consistency. For instance, DiffVideoAE~\cite{kim2023diffusion} achieves fine-grained editing of face-based speech videos by modifying face attributes or utilizing CLIP signals. Tune-a-Video~\cite{wu2022tune} inflates the image diffusion model and finetunes only on the given video to enable text-based editing. Pix2Video~\cite{ceylan2023pix2video} on the other hand, achieves training-free and consistent text-edited videos by injecting self-attention features from the previous frame into the current frame, implicitly aggregating temporal information. Layered neural representation \cite{kasten2021layered,lu2020layered} is another promising video editing method that aims to decompose video into different layers. Text2Live~\cite{bar2022text2live} combines such a representation with text guidance to show compelling video editing results.

With the ongoing advancements in generative AI techniques, a multitude of video generation platforms have surfaced. One notable example is the renowned Pika platform\footnote{https://github.com/pika/pika}, which serves as an idea-to-video platform, leveraging AI to create and edit videos seamlessly.

\noindent \textbf{Video Prediction} refers to the task of predicting future frames in a video sequence based on the observed past frames. Video prediction tasks have broad social implications, enhancing entertainment, improving security, aiding in understanding human behavior, and advancing autonomous systems. For example, it can be deployed to autonomous systems to plan and navigate their environment more effectively. Early recurrent-based works like FRNN~\cite{oliu2018folded} functionalize by recurrently inputting previous predictions to generate subsequent frames. To address the fact that RNNs tend to lead to blurry results, Hier-vRNN~\cite{castrejon2019improved} increases the expressiveness of the latent distributions using a hierarchy of latent variables. Most recently, conditional diffusion models also exhibit impressive results in video prediction. By conditioning on previous frames, RaMViD~\cite{hoppe2022diffusion} incorporates random conditioning masking to enable diffusion models to simultaneously perform prediction, infilling, and prediction tasks. MVCD~\cite{voleti2022mcvd} also finds that randomly and independently making out all the past frames or all the future frames in the training tends to generate high-quality predicted frames. FDM~\cite{harvey2022flexible}, on the other hand, found that selective sparse and long-range conditioning on previous frames is effective for generating long videos.

\subsection{Video Scene Understanding}
\begin{figure}[t!]
    \centering
    \includegraphics[width=\columnwidth]{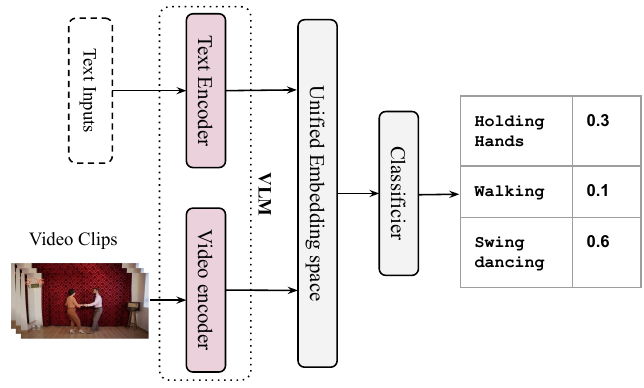}
    \caption{An illustration of VLMs for human action recognition. The input example is taken from the kinetics human action video dataset~\cite{kay2017kinetics}.}
    \label{fig:action-recog}
\end{figure}

\noindent \textbf{Human Action and Behavior Recognition} is one of the core tasks in video scene understanding, which aims to estimate human motion and behavior for online videos~\cite{surek2023video,hu2022online,morshed2023human}. In this context, it is required to analyze the motion and behavior considering the diversity of human body sizes, postures, view directions, lighting conditions, and camera movements, \textit{etc.} For this task, the major challenge is how to leverage the pre-trained LLMs to learn a strong representation of human motion from the video sequences~\cite{zhang2023large}. LLMs have been recently applied to diverse human action and recognition tasks. An illustration of LLM-guided action recognition is shown in Fig.~\ref{fig:action-recog}. For example, Kaneko \textit{et al.}~\cite{kaneko2023toward} proposed a method using LLMs to obtain new features for human activities based on the text prompt design. Zhou \textit{et al.}~\cite{zhou2023tent} proposed an approach to connect the signals, such as camera video, Lidar, and mmWave from the internet-of-things (IoT) sensors with LLMs to achieve the goal of human action recognition. By aligning the visual and language representation space, it is possible to directly map the visual features with the linguistic features. As such, the learned models are equipped with the zero-shot learning capacity to recognize unseen objects by imitating how humans recognize the objects. Wu \textit{et al.} \cite{wu2023bidirectional} introduced a video-text recognition framework that uses the natural language of vision-language models (VLMs), such as CLIP~\cite{radford2021learning} to bridge the video domain for cross-modal knowledge extraction.

\begin{table*}[t!]
\centering
\caption{The representative methods for video scene understanding.}
\begin{tabular}{l|l|l}
\hline
Method &Input modalities & Highlight \\
\hline
 \multicolumn{2}{c}{\textbf{\textit{Human Action and Behavior Recognition}}} & \\
\hline
Kaneko \textit{et al.}~\cite{kaneko2023toward} & Text, video & Designing text prompts to obtain new features. \\
Zhou \textit{et al.}~\cite{zhou2023tent} & Text, video, Lidar, nnMave & Aligning visual and language representation space for human action recognition. \\
Kaneko \textit{et al.}~\cite{kaneko2023toward} & Video, text & Using VLMs to bridge video domain for cross-modal knowledge extraction. \\
\hline
\multicolumn{2}{c}{\textbf{\textit{Video-based Dialogue and Conversation}}} &  \\
\hline
Video-ChatGPT~\cite{Maaz2023VideoChatGPT}  & Text, video & Capturing spatial-temporal relationships between video frames with LLM. \\
VideoChat~\cite{li2023videochat}  & Text, video & Video-centric dialogue system based on video foundation models and LLM. \\
Liu \textit{et al.}~\cite{liu2023one}  & Text, video &temporal modeling for video conversation tasks. \\
\hline
\multicolumn{2}{c}{\textbf{\textit{Human-Robot/Machine Interaction}}} &  \\
\hline
PaLM-E~\cite{driess2023palm} & Text, image, video &  A single large embodied multimodal model to tackle diverse embodied reasoning tasks. \\
LM-Nav~\cite{shah2023lm}   & Text, video & A robot-dialogue system for seamless interaction with humans based on video inputs. \\
\hline
\end{tabular} 
\label{reviewed_understand_methods}
\end{table*}

With LLMs or VLMs as guidance, human action and object recognition methods have been widely applied to video surveillance~\cite{wu2022survey}, robotic navigation~\cite{zhang2023large,singh2023progprompt,brohan2023rt}, medical diagnosis and healthcare~\cite{deng2022problem}, sports~\cite{wu2022survey}. For instance, LLMs with vision sensors enable robots with stronger NLP capacity based on the video sequences. This enables more intensive integration between the human and robot by imitating human reasoning and conversations. In sports,  the zero-shot recognition capacity and semantic richness of LLMs are used to guide the action recognition models for diverse sports activities, such as football and basketball.

In summary, the fusion of LLMs with videos for human action and object recognition heralds
an exciting epoch for video scene understanding. With active research being made, it enjoys a great benefit for a broader range of video-based applications. 

\begin{figure}[t!]
    \centering
    \includegraphics[width=\columnwidth]{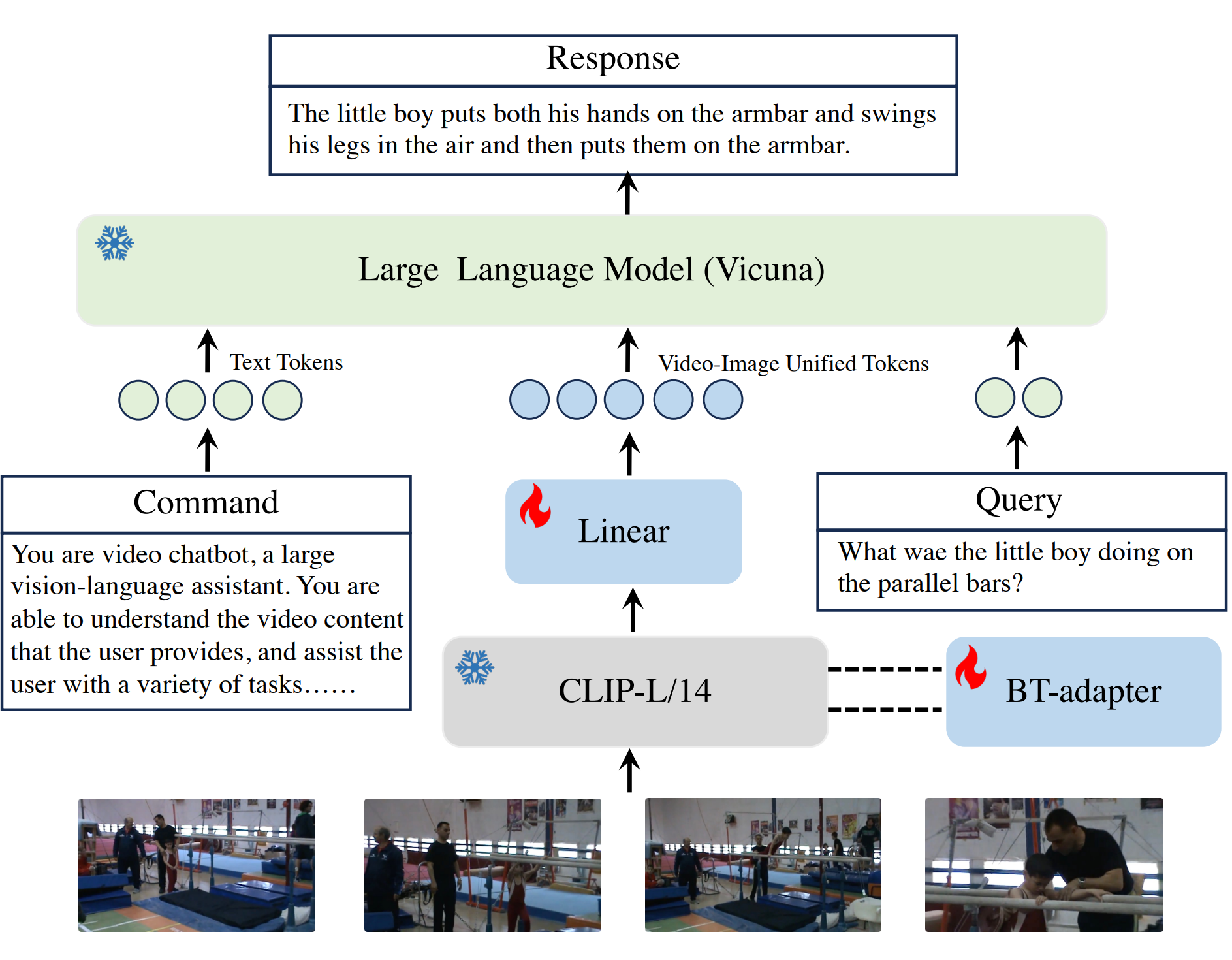}
    \caption{A representative pipeline of video conversation based on LLMs~\cite{liu2023one}. }
    \label{fig:dialogue-conv}
\end{figure}

 \noindent \textbf{Video-based Dialogue and Conversation.}
LLMs are able to provide semantic information and generate symbolic spatial signals, which can serve as guidance for video scene understanding. Recently this has been demonstrated for interactive video-based dialogue and conversation~\cite{maaz2023video,liu2023one,li2023videochat,luo2023valley,zhang2023video,li2023unmasked}. In this context, Video-ChatGPT \cite{maaz2023video} is designed for video understanding and conversation by capturing the spatial-temporal relationships between video frames based on LLMs. It demonstrates strong conversation and contextual understanding capabilities on diverse benchmark datasets. VideoChat~\cite{li2023videochat}, on the other hand, introduces a video-centric multi-modal dialogue system that integrates the video foundation models and LLMs. Moreover, Liu \textit{et al.}~\cite{liu2023one} extended the LLMs to the video domain and incorporated a spatial-temporal module for temporal modeling for the video conversation tasks, as depicted in Fig.~\ref{fig:dialogue-conv}. 

To summarize, recent progress in video-based dialogue and conversation has been primarily demonstrated by the integration of video/image-based models with LLMs. With LLMs, it is possible to achieve zero-shot conservation by exploring the temporal relationship with video-centered dialogue modeling.

\begin{figure}[t!]
    \centering
    \includegraphics[width=\columnwidth]{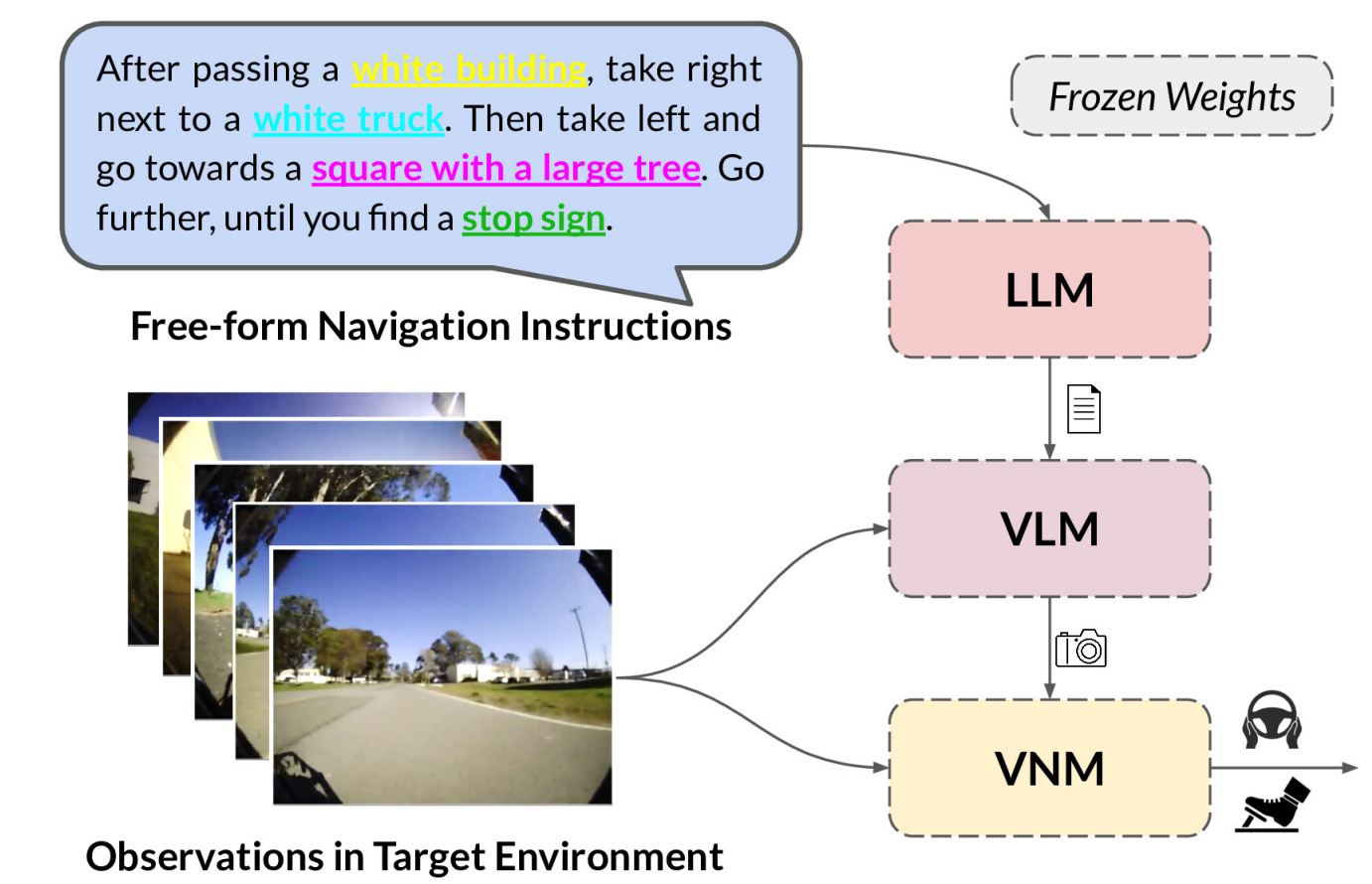}
    \caption{An example of navigation instructions based on the large language model (LLM) for landmark extraction, a vision-and-language model (VLM) for grounding, and a visual navigation model (VNM) for execution~\cite{shah2023lm}. }
    \label{fig:llm-hri}
\end{figure}

\noindent \textbf{Human-Robot/Machine Interaction.}
With the popularity of LLMs, many research endeavors have been devoted to the application of LLMs in the field of human-robot/machine interaction, as exemplified by the visual illustration in Fig.~\ref{fig:llm-hri}. On one hand, with pre-trained LLMs, robots are endowed with the capacity to understand human needs and queries~\cite{driess2023palm}. On the other hand, LLMs enable robots to articulate fluent and human-like natural language via interaction with LLMs~\cite{shah2023lm}. However, applying  LLMs for human-robot/machine interaction needs to deal with the inaccurate reasoning provided by the LLMs. To this end, robot-dialogue systems are developed for more seamless interaction with humans based on camera video inputs.

As an emerging area, this direction exhibits great potential and it provides new paradigms for robot navigation and human-robot interaction. LLMs help enhance learning efficiency and performance, and meanwhile, strengthen the interaction between humans and robots. 

\subsection{Streaming}

While the use of LLMs in video streaming is still in its infancy, the potential applications in areas such as user viewing angle prediction, network condition prediction, and video content encoding and processing suggest significant development opportunities. Ongoing research and innovation are poised to propel the application of LLMs in video streaming, ultimately offering users more intelligent and personalized viewing experiences.
In this context, we delve into several classic applications of transformer-based LLMs within the realm of video streaming.

\noindent \textbf{360° and Volumetric Video Streaming.} 360° in general is one spherical video stitching multiple videos recorded by a set of cameras or lenses filming different angles of a view simultaneously. Once the videos are merged into one, the different shots are synchronized in terms of color and contrast by either the camera or video editing software. 
In order to compress the 360° videos using a standard codec (such as H.264~\cite{wiegand2003overview} and HEVC~\cite{sullivan2012overview}), the video is projected into 2D domain. 360°  video is much larger (4$\times$ to 6$\times$) than conventional videos under the same perceived quality due to their panoramic nature. The ultimate 360°  video with single-eye 8K resolution requires the bandwidth to reach multiple Gigabits-per-second (Gbps), posing a great challenge on the network and a huge burden on the cost \cite{huawei17,ai2022deep}. The mainstream industry believes that the motion-to-photons latency (MTP) should not exceed 20 ms\footnote{Huawei-iLab. 2018. Cloud VR Network Solution White Paper. Retrieved from http://www.huawei.com/}, or otherwise would cause dizziness to users.

Volumetric video (or hologram video), the medium for representing natural content in VR/ AR/MR, is presumably the next generation of video technology and a typical use case for 5G and beyond wireless communications \cite{van2020capturing,liu2021point}. Volumetric video provides users with six degrees of freedom (6DoF) immersive viewing experience, that is, users can freely move forward/backward (surging), up/down (heaving), or left/right (swaying) to select their favorite viewing angle of the 3D scene, and hence enjoy three more degrees of freedom in comparison with 3DoF VR video users. As the most popular and favored representation of volumetric media, point clouds consist of 3D points, each with multiple attributes, such as coordinates and color. 

For both 360° and volumetric videos, each time, a user perceives part of the 360° scene, namely field-of-view (FoV). As the user rotates his/her head, correspondingly different FoV of the 360° scene is rendered for observation. By allowing users to freely select any viewing angles inside the
video sphere, 360° and volumetric videos bring the immersive viewing experience to a new
level compared to traditional video and multi-view video. 


Compared with traditional video streaming, the technical challenges of 360° and volumetric videos include:
\begin{itemize}
\item Viewport prediction: Each user each time only observes a portion of the 360° scene and may switch FoVs during the video playback. Also, addressing inevitable wrong viewpoint predictions is important to guarantee the quality of video services.
\item Strict latency requirement: MTP needs to be under 20 ms.
\item Tiling-based resource allocation: 360° and volumetric video streaming resource allocation is conducted at the tile level and has to consider the quality switches. 
\end{itemize}

The technologies backed by LLMs mentioned in Section~\ref{subsec:streaming}, including viewport prediction, bandwidth prediction, compression, and resource allocation can jointly optimize the challenging streaming tasks for 360° and volumetric Videos.


\noindent\textbf{Short Video Recommendation.} 
Short Videos have become increasingly popular in recent years, with platforms such as TikTok and YouTube Shorts providing platforms for users to create and share content. These videos typically range from a few seconds to a minute in length and cover a wide range of topics. The rise of short videos has revolutionized the way we consume and create content, making it easier than ever for anyone to share their ideas and creativity with the world. 

From a technical standpoint, the transmission of these videos is quite different from that of regular videos \cite{guo2021video}. Typically, servers recommend a set of videos to the user (e.g. 5), and all of these videos are pushed to the user. The user then selects which videos to watch and discards the ones they don't like, resulting in wasted transmission resources. However, if not all the videos are transmitted, the user may experience buffering or a decrease in video quality, which can significantly impact their viewing experience. 
This issue involves how to recommend videos to the user, whether to transmit all or part of the videos, and how to allocate video resources, among other challenges. Furthermore, there is a lack of available video libraries for research, which presents a significant obstacle.
Accurate recommendations are crucial to minimize the waste of bandwidth. Video recommendation systems incorporated with LLMs can better comprehend user preferences and context, leading to more accurate and personalized video recommendations. LLMs can analyze user queries, video descriptions, and other textual information associated with videos to grasp the semantic meaning, sentiment, and other important factors that impact the user's preferences. This approach has the potential to significantly enhance user satisfaction, engagement, and retention within video platforms. As these language models continuously learn from vast amounts of textual data, they become increasingly adept at understanding user intent and preferences, resulting in more relevant and appealing video recommendations. Ultimately, this improvement in video recommendation can lead to a more enjoyable and immersive user experience, benefiting both users and video content providers.


\begin{figure}[tb!]
	\centering
	\includegraphics[width=0.48\textwidth]{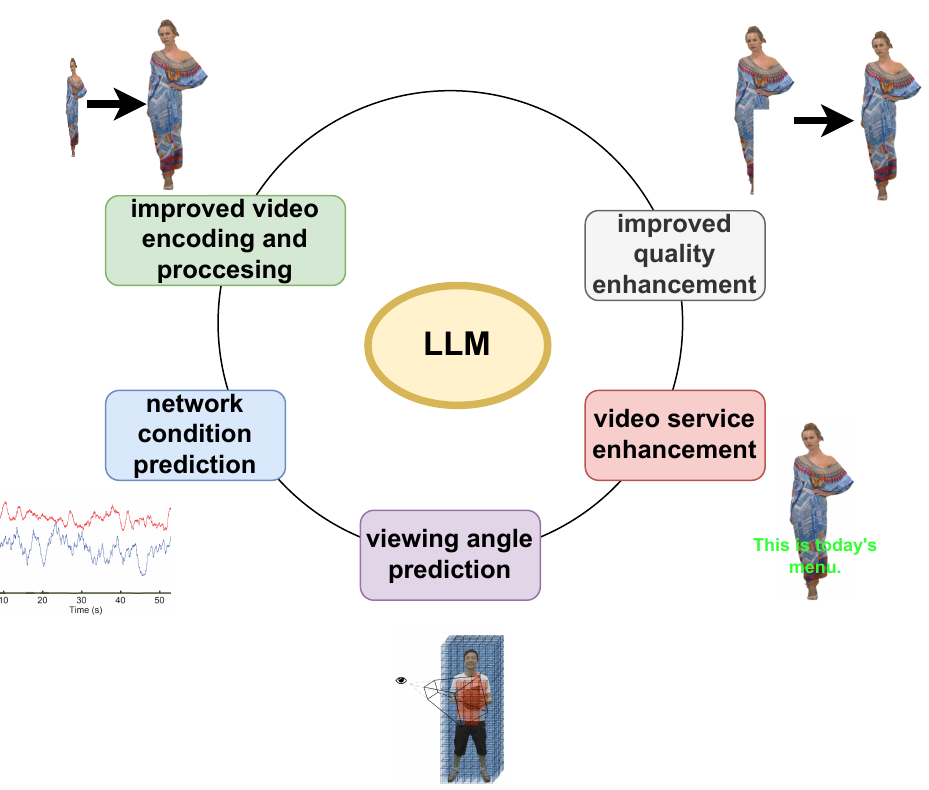}
	\caption{Illustration of the application of LLM in video streaming.}
	\label{fig:llmvideoapp}
\end{figure}

\noindent\textbf{Video Service Enhancement.} Transformer-based LLMs can be applied to image super-resolution, enhancing video quality by predicting and generating higher-resolution images, or removing artifacts from a lossy compressed video and the improvement of the visual properties by a photo-realistic restoration of the video contents. For instance, Liu et al. introduced a pioneering trajectory-aware Transformer in \cite{liu2022learning}, marking one of the initial attempts to integrate Transformer architectures into video super-resolution tasks. The proposed model demonstrates excellent performance. Geng et al. presented a unified spatial-temporal transformer that integrates temporal interpolation and spatial super-resolution modules for space-time video super-resolution \cite{geng2022rstt}. This innovative approach results in a significantly smaller network compared to existing methods, enabling real-time inference without substantial performance compromise. 
\cite{vasluianu2022efficient} introduced a real-time online video enhancement transformer characterized by low latency, utilizing spatial and temporal attention mechanisms. The proposed model demonstrates quantitative and qualitative advancements over state-of-the-art methods with minimal inference time.

Video service enhancement with LLMs and Generative AI has also shown notable advancements recently. \cite{renella2023towards} presented an innovative approach to automatically generate streaming commentary during the game \textit{League of Legends}. This system adeptly identifies key events and utilizes generative AI services to craft voice output. Additionally, \cite{lin2022swinbert} introduced a comprehensive transformer-based model for video captioning, an important service in streaming. The authors propose the sparse attention mask as a regularization technique to improve long-range video sequence modeling. They also provide quantitative validation, affirming the efficacy of the learnable sparse attention mask in the realm of caption generation.

\begin{table*}[t! ]
\centering
\caption{The reviewed LLM methods for video streaming.}
\begin{tabular}{l|l|l}
\hline
Method &Input information &Task \\
\hline
\multicolumn{2}{c}{\textbf{\textit{Viewport prediction}}} & \\
\hline
transformer-based approach~\cite{chao2021transformer}  & past viewing angle scanpaths  & long-term viewing angle predictions \\
 &   &  with low complexity. \\

transformers-based approach~\cite{cheng2022gaze}  & face images & eye-gaze information.\\
spatio-temporal transformer~\cite{ni2023human}  & gaze features, scene contexts and the & forecast human–object interactions in videos \\& visual characteristics of human–object pairs.& \\
\hline
\multicolumn{2}{c}{\textbf{\textit{Bandwidth prediction}}} & \\
\hline
transformer-based model~\cite{azmin2022bandwidth} & previous bandwidth info. & future bandwidth condition. \\
GAN-based solution~\cite{kattadige2021videotrain} & actual video traces  & synthesizing video streaming data, with a  \\
 &   & focus on 360°/normal video classification. \\
\hline

\multicolumn{2}{c}{\textbf{\textit{Video compression}}} & \\
\hline
masked image modeling transformer ~\cite{xiang2022mimt} &  video  & deep video compression. \\
 transformer-based approach  \cite{mentzer2022vct} & video  & neural video compression. \\
\hline
\multicolumn{2}{c}{\textbf{\textit{Video enhancement}}} & \\
\hline
video Enhancement transformer ~\cite{vasluianu2022efficient} &  original video  & video with enhanced quality. \\
transformer-based method~\cite{liu2022learning} &  video  & video super-resolution. \\
unified spatial-temporal transformer ~\cite{geng2022rstt} & video  & space-time video super-resolution.  \\
GAN model \cite{angarano2023generative} & video  & real-time super-resolution. \\
transformer-based model  ~\cite{lin2022swinbert} &  video to be watched& video captioning. \\
\hline
\hline
\end{tabular} 
\label{reviewed_stream_methods}
\end{table*}

\section{Challenges}
In this section, we discuss the major challenges faced by Generative AI and LLMs when employed for video generation, understanding, and streaming services.
\subsection{Generation}
\noindent \textbf{Temporal Consistency.} One of the main challenges in Generative AI for video content creation is ensuring temporal consistency between the generated frames. Generated video sequences should exhibit smooth and realistic motion patterns, and maintaining these patterns across frames can be challenging for generative models. In addition to the amount of video, the training strategy choice also plays a pivotal role in terms of consistency. Modeling the video generation as a discrete image generation task will easily lead to poor temporal consistency and suffer from temporal flickering~\cite{tian2021good,villegas2017decomposing}. Implicit neural representations (INRs) based methods \cite{yu2022generating} by treating the time axis as a continual signal could be easily deployed to generate arbitrary long videos. TGANv2~\cite{saito2020train} addresses the problem by introducing a hierarchical discriminator to guarantee smoothness in the levels from coarse to fine. Recent image pretrained models \cite{blattmann2023align} find that interplacing multiple temporal attention layers and fully finetuned on video datasets is another effective way. 

\noindent \textbf{High Computational Requirements.} Video generation requires processing high-dimensional data, which significantly increases the computational requirements for training and inference. Developing efficient algorithms and parallelization techniques for video generation remains an ongoing challenge. Works like NUWA~\cite{wu2022nuwa} and Imagen-Video~\cite{ho2022imagen}, which belong to the text-video generator category, are trained on millions of text-video pairs, making them challenging to replicate for most research groups. However, certain editing-based video generation approaches address the computational burden by utilizing a small amount of video dataset or even none at all to achieve specific tasks. Tune-a-Video~\cite{wu2022tune} is an example of such a method, where fine-tuning a video is accomplished by leveraging an image generator to accomplish targeted editing tasks. These specific task-driven videos, due to their constrained sample space and lower requirement for model temporal modeling capabilities, constitute a direction that can be widely explored.

\noindent \textbf{Lack Large-Scale Video Datasets.} While large-scale image datasets are widely available, video datasets of similar scale and diversity are scarce. The lack of large-scale video datasets hinders the development of Generative AI models for video content creation, as they rely on large amounts of data to learn the underlying data distribution. Annotated video datasets are relatively scarce, yet they play a crucial role in controllable video generation. Due to the highly redundant nature of video content, some recent studies \cite{ho2022imagen,skorokhodov2022stylegan,blattmann2023align} have leveraged powerful pretrained text-image generators to initialize the spatial modeling network layers, resulting in improved quality in a single-frame generation. This allows the temporal modules to focus more on modeling the dynamics of the sequential signals. Additionally, certain approaches \cite{ho2022video,voleti2022mcvd} have addressed the data scarcity issue by employing image-video joint training techniques, which exhibit a trade-off between temporal consistency and frame fidelity at the same time.

\subsection{Understanding}

\noindent \textbf{Temporal Reasoning.} Video scene understanding involves reasoning over temporal information, including the dynamics, actions, and interactions within a video. However, LLMs often struggle with effectively capturing and modeling long-range temporal dependencies. Temporal reasoning in videos is challenging due to the varying lengths of videos and the need to recognize and contextualize actions over time. Developing LLM architectures that can effectively reason over long-term dependencies, capture temporal context, and understand the dynamics of video scenes is a significant research challenge. Techniques such as temporal convolutions, recurrent neural networks, or attention mechanisms need to be explored to improve the temporal reasoning capabilities of LLMs.

\noindent \textbf{Multimodal Understanding.}  Videos consist of both visual and audio information, and understanding videos comprehensively requires multimodal understanding~\cite{zhao2023learning}. LLMs need to effectively integrate visual and auditory modalities to capture the full context and meaning of video scenes. However, aligning and connecting the visual and audio information in videos is an intricate task. Therefore, it is imperative to explore network architectures and methods for effectively modeling audio-visual interactions, capturing the cross-modal dependencies, and fusing multimodal information in LLMs~\cite{guo2023evaluating}. Moreover, developing methods for training LLMs on large-scale multimodal video datasets that cover a wide range of scenes and languages is crucial for enhancing their multimodal understanding capabilities.

\noindent \textbf{Real-Time Video Processing.}  Processing videos together with LLMs in real-time poses a significant challenge. Real-time video scene understanding is crucial for various applications such as autonomous vehicles, surveillance systems, and video analytics~\cite{huang2022towards}. However, the large model size and computational requirements of LLMs hinder their real-time processing capabilities. Therefore, further research is required to develop efficient networks, model compression approaches, and hardware optimizations to accelerate the inference of LLMs for video scene understanding. Techniques, such as knowledge distillation~\cite{wang2021knowledge,zhu2023good}, pruning, and quantization can be explored to reduce the computational burden and enable real-time processing of videos with LLMs. Furthermore, exploring distributed computing and hardware accelerators can further enhance the real-time capabilities of LLMs for video scene understanding~\cite{guo2023evaluating}.

\noindent \textbf{Limited Performance of Zero-shot.} Although LLMs deliver exceptional zero-shot learning capacity, however, it is hardly possible to enable the LLM-guided video scene understanding models to have the same capacity. Similar to video generation, the major challenge is the lack of large-scale paired video-text datasets due to the difficulty of generating rich textual descriptions for the video clips. Thus, it is difficult to learn strong representations for the target tasks. Another reason is that, for the long-form videos, the text annotations are either sparse or not sufficient to illustrate the happening event or activities. Therefore, future research might explore how to leverage LLMs to impose more effective supervision given the limited or sparse text descriptions.  Another direction is how to leverage LLMs to further generate high-quality video-text pairs with more semantic richness.

\subsection{Streaming}
\noindent \textbf{Varied Environments and Demands.}
Considerable variations exist in the computational capabilities, resolutions, and network conditions of devices used by users to watch videos. Additionally, diverse video transmission ways (such as live streaming and video-on-demand) and video types (such as VR videos and short videos) impose varying bandwidth, experimental, and computational requirements on transmission. Designing or learning an algorithm to adapt to these heterogeneous scenarios is a formidable task. LLMs have the capacity to encompass these situations and provide solutions to the problem. However, when employing LLMs for video transmission scheduling, effectively addressing these challenges and providing answers within a short timeframe (given the strong demands of video on algorithm complexity) is a non-trivial and substantial challenge, necessitating further research in the future.

\noindent \textbf{A Unified Framework or Standard.}
Traditional video transmission methods have reached a high level of maturity, giving rise to widely used applications like YouTube and Zoom. A significant contributing factor in this domain is the introduction of the MPEG-DASH video transmission standard \cite{sodagar2011mpeg}, which laid the foundation for video transmission strategies. Companies and research groups have since been able to innovate and establish new applications based on this framework. However, there is currently no unified video transmission framework or standard in the context of LLM-based video transmission. Divergent technical approaches hinder the development of this field. Establishing a unified video transmission framework or standard is a challenging task, requiring the participation of numerous entities.

\noindent \textbf{Lack of Large-Scale Video Datasets.}
Similar to the preceding discussions on generation and understanding, when leveraging LLM for optimization and scheduling in the realm of transmission, learning is imperative. This naturally leads to the need for datasets. Presently, there are publicly available datasets for individual aspects such as network bandwidth \cite{vanderHooft2016}, video data, and user data for VR videos, such as those provided by MPEG \footnote{https://www.mpeg.org/standards/}. However, in comparison to the requirements for LLM learning, these datasets are relatively small, and datasets possessed by major corporations are not open-source. Furthermore, comprehensive datasets annotating communication states, user devices, user viewing data, user satisfaction, etc., are currently lacking. Generative AI may contribute to generating datasets for training models for bandwidth prediction.  \cite{kattadige2021videotrain} introduced an innovative GAN solution for synthesizing video streaming data, with a focus on 360°/normal video classification. This approach demonstrated an improvement in accuracy compared to relying solely on actual traces.

\section{Concerns}
Aside from attractive potentials, Generative AI and LLMs also raise considerable concerns that should be addressed properly. Noticeable concerns include misleading information dissemination via video forgery and intellectual property rights violations, among others.
\\

\begin{figure}[t!]
	\centering
	\includegraphics[width=\columnwidth]{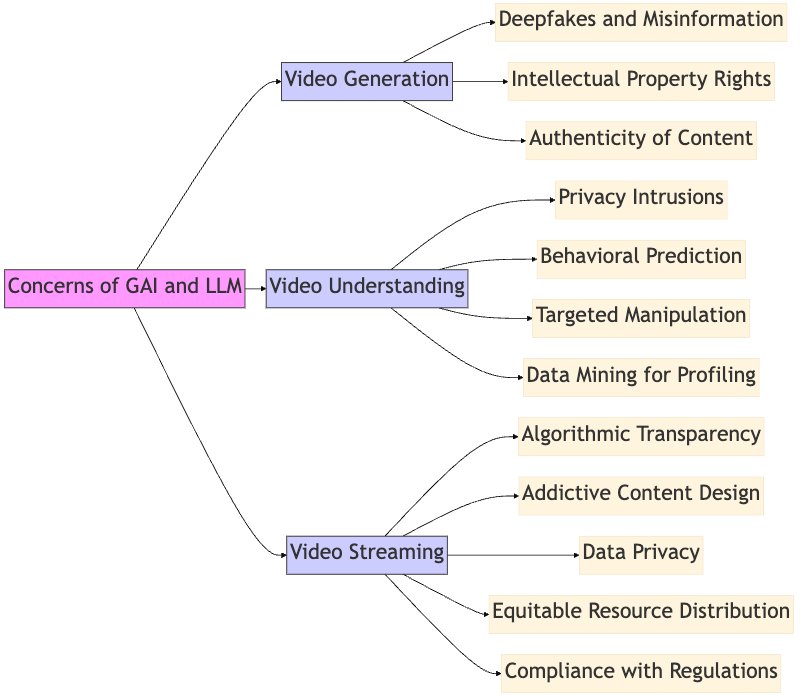}
	\caption{Concerns faced by GAI- and LLM- based solutions.}
	\label{fig:concerns}
\end{figure}

\noindent\textbf{Misinformation} The improving GAI's ability to generate seemingly authentic video footage can be misused for creating false narratives, propagating fake news, impersonating individuals without their consent, or manipulating public opinion, resulting in severe impacts on society in terms of politics, security, and trustworthiness. The increasing number of reported events in this direction has raised wide corners from the society~\footnote{\url{https://www.nbcnews.com/tech/tech-news/deepfake-scams-arrived-fake-videos-spread-facebook-tiktok-youtube-rcna101415}}.

\noindent\textbf{Intellectual property right violation.} Generative AI has been continuously improved to edit and revise the style and details of existing videos, infringing copyright and using proprietary content without authorization.

\noindent\textbf{Security.} Generative AI can make deepfake videos to mimic legitimate videos from trusted sources or individuals, facilitating fraud and cybercrime. There has also been an increasing number of relevant cases reported in recent years~\footnote{\url{https://www.bbc.com/news/technology-66993651}}.

\noindent\textbf{Privacy leakage.} LLMs, if employed in already-everywhere surveillance systems, can not only identify individuals but also infer their activities and routines. This could lead to a serious privacy concern where people are constantly monitored, violating the right to privacy. Further, when deployed with monitors equipped with audio receivers, LLM can potentially eavesdrop on private conversations.

\noindent\textbf{Content censorship.} LLM-driven streaming services, while providing the potential to improve user experience, can also result in the over-filtering of the content, which might amount to censorship. Determining what content reaches the audience without clear guidelines can lead to arbitrary content suppression.

\noindent\textbf{Bias.} The existing bias issues such as stereotypes could worsen with the use of Generative AI and LLM. Personalized streaming recommendations can reinforce existing biases and isolate users from diverse perspectives. The risk also applies to the generation stage of videos.

\noindent\textbf{Addictive content design.} Generative AI can be used to generate certain types of videos optimized for maximum engagement, potentially leading to addictive content exploiting human psychology to increase screen time.
\\

Overall, integrating Generative AI and LLM into the video industries introduces a multitude of concerns that span privacy, ethics, and societal impact, among others. In video generation, the ability to create hyper-realistic deepfakes poses significant risks for misinformation, privacy violations, and intellectual property infringements. The improving understanding capabilities of LLMs on videos raise alarms about privacy intrusions, such as sensitive data mining for personalized profiling and behavioral prediction that could be exploited for targeted manipulation. In streaming, opaque recommendation systems can create content bubbles and potentially skew the cultural narrative. Additionally, the personalization of content raises ethical concerns about data privacy, the psychological impact of addictive content designs, and equitable resource distribution.

To address these concerns, proactive and cautious actions are required. Regulators should craft robust privacy protections and transparency mandates that compel video services to disclose how user data informs content delivery. Ethical AI frameworks should be set up to guide the creation and use of video service algorithms to avoid bias and make sure that the content available is diverse and fair. Video platforms must prioritize user consent and data security by implementing best practices for data handling and providing users with clear choices regarding their data. There's also a need for an industry-wide commitment to ethical content design, avoiding manipulative practices, and promoting mental well-being. Finally, video services must ensure compliance with international regulations through adaptive AI systems that can meet local standards while respecting global norms. Through these concerted efforts, the industry can harness the benefits of AI for video services while safeguarding individual rights and societal values.
\section{Conclusion}
In this paper, we conduct a comprehensive examination of how generative artificial intelligence (Generative AI) and large language models (LLMs) are revolutionizing the video technology sector, focusing on video generation, understanding, and streaming. The innovative integration of these technologies results in highly realistic digital creation, enhanced video understanding by extracting meaningful information from visual content, and more efficient and personalized streaming experiences, thus improving user interaction with videos and user preference-tailored experience provision.

The paper navigates through current achievements, ongoing challenges, and future possibilities in applying Generative AI and LLMs to video-related tasks. It underscores the immense potential these technologies hold for advancing video technology across multimedia, networking, and AI communities. It also highlights the challenges and concerns that require further exploration.

Observed from the reviewed works, we can see that, overall, advanced AI technologies like GAI and LLMs are making profound impacts on several key sectors of video-related research fields. The biggest advantage of AI-based methods is their automation capability with lower manual costs. However, it comes at the price of challenges uniquely faced by AI, such as lack of large-scale datasets, high computational cost, consistency issues, and concerns such as misinformation and security, etc. Therefore, academia and industry should be cautious during the rapid development to ensure a sustainable market. 
\bibliographystyle{IEEEtran} %
\bibliography{ref}

\end{document}